\documentclass[11pt]{article}

\usepackage[utf8]{inputenc}
\usepackage[T1]{fontenc}
\usepackage{lmodern}
\usepackage{microtype}
\usepackage[a4paper,margin=1in]{geometry}
\usepackage{amsmath,amssymb}
\usepackage{cite}
\usepackage{booktabs}
\usepackage{longtable}
\usepackage{array}
\usepackage{tabularx}
\usepackage{graphicx}
\usepackage{caption}
\usepackage{float}
\usepackage[section]{placeins}
\usepackage{needspace}
\usepackage{xurl}
\usepackage{url}
\usepackage{hyperref}
\usepackage{enumitem}
\usepackage[most]{tcolorbox}
\usepackage{xcolor}

\hypersetup{
    colorlinks=true,
    linkcolor=blue,
    citecolor=blue,
    urlcolor=blue,
    pdftitle={QuantCode-Bench},
    pdfauthor={Alexey Khoroshilov, Alexey Chernysh, Orkhan Ekhtibarov, Nini Kamkia, Dmitry Zmitrovich}
}

\newtcolorbox{taskbox}{
    enhanced,
    breakable,
    colback=gray!4,
    colframe=black!70,
    boxrule=0.8pt,
    arc=1mm,
    left=2mm,
    right=2mm,
    top=1.5mm,
    bottom=1.5mm
}

\setlist[itemize]{leftmargin=1.5em}
\setlist[enumerate]{leftmargin=1.8em}

\title{QuantCode-Bench: A Benchmark for Evaluating the Ability of Large Language Models to Generate Executable Algorithmic Trading Strategies}

\author{
\begin{tabular}{ccc}
Alexey Khoroshilov & Alexey Chernysh & Orkhan Ekhtibarov\\[2pt]
\multicolumn{3}{c}{Nini Kamkia \qquad Dmitry Zmitrovich}
\end{tabular}\\[6pt]
Lime
}

\date{}

\begin{document}

\maketitle

\begin{abstract}
Large language models have demonstrated strong performance on general-purpose programming tasks, yet their ability to generate executable algorithmic trading strategies remains underexplored. Unlike standard code benchmarks, trading-strategy generation requires simultaneous mastery of domain-specific financial logic, knowledge of a specialized API, and the ability to produce code that is not only syntactically correct but also leads to actual trades on historical data. In this work, we present \textbf{QuantCode-Bench}, a benchmark for the systematic evaluation of modern LLMs in generating strategies for the \textbf{Backtrader} framework from textual descriptions in English. The benchmark contains \textbf{400 tasks} of varying difficulty collected from Reddit, TradingView, StackExchange, GitHub, and synthetic sources. Evaluation is conducted through a multi-stage pipeline that checks syntactic correctness, successful backtest execution, the presence of trades, and semantic alignment with the task description using an LLM judge. We compare state-of-the-art models in two settings: \textbf{single-turn}, where the strategy must be generated correctly on the first attempt, and \textbf{agentic multi-turn}, where the model receives iterative feedback and may repair its errors. Results show that even the best frontier models achieve only about \textbf{70--76\%} Judge Pass in the single-turn setting, whereas the best models in the agentic setting reach \textbf{95--98\%}. We analyze the failure modes across different stages of the pipeline and show that the main limitations of current models are not related to syntax, but rather to the correct operationalization of trading logic, proper API usage, and adherence to task semantics. These findings suggest that trading strategy generation constitutes a distinct class of domain-specific code generation tasks in which success requires not only technical correctness, but also alignment between natural-language descriptions, financial logic, and the observable behavior of the strategy on data.
\end{abstract}

\section{Introduction}

The ability of large language models to generate code has rapidly become one of the central areas for evaluating modern AI systems. However, most existing benchmarks focus either on general programming tasks, code repair, or repository-level software engineering problems. Such benchmarks include, for example, \textbf{SWE-Bench Verified}, \textbf{LiveCodeBench}, \textbf{Terminal-Bench 2.0}, and \textbf{SWE-rebench}, which have substantially advanced the evaluation of LLMs in code repair, self-repair, and agentic software engineering~\cite{jimenez2023swe,jain2024livecodebench,badertdinov2025swerebench,merrill2026terminalbench}. Nevertheless, these benchmarks do not fully capture model behavior in domain-specific applied settings, where a model must simultaneously understand the subject matter, follow a specialized API, and produce code that exhibits meaningful behavior when executed.

One such task is the generation of algorithmic trading strategies. A model must interpret a textual description of a trading idea, identify the indicators, entry and exit conditions, position management rules, and possible parameters. It must then translate this description into correct code for a specific framework, in our case \textbf{Backtrader}, while respecting its interfaces, indexing conventions, indicator syntax, and execution semantics~\cite{backtrader}. Finally, the generated code must not only execute successfully, but also produce actual trading signals on historical data. Even when syntax is fully correct, a strategy may still be non-functional because of overly strict thresholds, incorrect interpretation of conditions, or the absence of a link between indicators and trading actions.

This combination of requirements makes trading strategy generation fundamentally different from most existing code benchmarks. In standard coding tasks, successful compilation and passing tests often serve as sufficient indicators of quality~\cite{jimenez2023swe,jain2024livecodebench}. In algorithmic trading, this is not enough. Code may be technically correct, may pass a backtest successfully, and still be functionally useless if it generates no trades. Moreover, a strategy may be executable and even place trades, yet still fail to match the original task description. As a result, this setting requires a stricter and more layered evaluation protocol.

Interest in applying LLMs to finance is growing rapidly, but most of the literature focuses on financial NLP, question answering, document analysis, information extraction, forecasting, and agentic retrieval. This is reflected in works such as \textbf{PIXIU}, \textbf{FinBen}, \textbf{FinanceBench}, \textbf{Fin-R1}, \textbf{Fino1}, \textbf{Finance Agent Benchmark}, and \textbf{FinAgentBench}~\cite{xie2023pixiu,xie2024finben,islam2023financebench,shi2025finr1,shah2025fino1,bigeard2025financeagent,choi2025finagentbench}. Against this background, there is still a lack of benchmarks that specifically measure the ability of LLMs to translate natural-language strategy descriptions into executable trading-system code.

In this work, we present \textbf{QuantCode-Bench}, a benchmark for evaluating the ability of LLMs to generate executable trading strategies from textual descriptions. The benchmark is built around the \textbf{Backtrader} framework and includes \textbf{400 tasks}. We consider two interaction settings. In the \textbf{single-turn} setting, the model must solve the task on the first attempt, without any opportunity for revision. In the \textbf{agentic multi-turn} setting, the model receives structured feedback after each failure and may iteratively improve the code. This setup makes it possible to separately evaluate a model's one-shot generation ability and its capacity to repair its own errors in an interactive scenario.

The core idea of \textbf{QuantCode-Bench} is that successful trading strategy generation should not be defined by a single criterion, but by a sequence of nested requirements. To this end, we use a four-stage evaluation pipeline: syntactic correctness, successful execution on historical data, the presence of at least one trade, and final semantic validation with an LLM judge. The use of \textbf{LLM-as-a-Judge} in open-ended tasks has already become widespread~\cite{zheng2023judge,gu2024surveyjudge}. This evaluation design makes it possible to distinguish at least four types of model capability: the ability to generate correct code, the ability to construct an executable strategy, the ability to formulate conditions that lead to real trading signals, and the ability to implement the actual trading idea described in the prompt.

Our experiments show that the task remains challenging even for the strongest models. In the single-turn setting, frontier models achieve nearly perfect compilation rates but degrade substantially at later stages of the pipeline. The best Judge Pass does not exceed roughly half of the benchmark. This means that the main challenge lies not in surface-level syntax, but in the deeper operationalization of trading logic. In the agentic setting, performance improves sharply: many errors prove to be locally repairable when feedback is available. However, a portion of the final failures in this setting still stem from incorrect interpretation of the natural-language specification rather than merely from local code defects.

Our contributions are as follows. First, we introduce a benchmark specifically designed for the generation of executable algorithmic trading strategies. Second, we propose a multi-level evaluation framework that distinguishes technical executability, the presence of trading behavior, and semantic alignment with the task specification. Third, we perform a broad comparison of modern models in both single-turn and agentic settings. Fourth, we conduct a detailed error analysis and identify the dominant failure modes at different stages of the pipeline. Fifth, we release the benchmark as a reproducible foundation for future research in domain-specific code generation for financial applications.

\section{QuantCode-Bench}

\subsection{Task Definition}

QuantCode-Bench evaluates a model's ability to generate a \textbf{Backtrader} trading strategy from a textual description, subject to four nested requirements. First, the strategy must be syntactically correct. Second, it must execute successfully within the backtesting environment. Third, it must place at least one trade on the provided historical data. Fourth, it must match the described trading idea rather than merely producing an arbitrary working template.

This formulation makes the benchmark substantially stricter than typical coding tasks. Each successive validation stage strengthens the notion of success. Successful compilation does not guarantee successful execution. Successful execution does not guarantee the presence of trading signals. The presence of trades does not guarantee compliance with the textual specification. Therefore, our primary overall metric is \textbf{Judge Pass}, defined as the proportion of tasks for which the generated strategy passes the entire evaluation pipeline.

\subsection{Dataset}

The \textbf{QuantCode-Bench} dataset contains \textbf{400 trading-strategy generation tasks}. The descriptions were collected from multiple sources that differ in formality, structure, and level of detail:

\begin{table}[ht]
\centering
\begin{tabular}{lr}
\toprule
\textbf{Source} & \textbf{Count} \\
\midrule
Reddit & 183 \\
TradingView & 100 \\
StackExchange & 90 \\
GitHub & 19 \\
Synthetic & 8 \\
\midrule
\textbf{Total} & \textbf{400} \\
\bottomrule
\end{tabular}
\caption{Source distribution of QuantCode-Bench tasks.}
\label{tab:sources}
\end{table}

Each task underwent structural enrichment. From the original description, we extracted the indicators used, the entry and exit conditions, and any additional rules, whether stated explicitly or implicitly. Each task was then assigned a difficulty category: \textbf{easy}, \textbf{medium}, or \textbf{hard}.

The distribution of tasks by source and difficulty is shown in Table~\ref{tab:difficulty}.

\begin{table}[ht]
\centering
\begin{tabular}{lrrrr}
\toprule
\textbf{Source} & \textbf{Easy} & \textbf{Medium} & \textbf{Hard} & \textbf{Total} \\
\midrule
Reddit & 147 & 16 & 20 & 183 \\
TradingView & 6 & 57 & 37 & 100 \\
StackExchange & 32 & 34 & 24 & 90 \\
GitHub & 12 & 1 & 6 & 19 \\
Synthetic & 0 & 8 & 0 & 8 \\
\midrule
\textbf{Total} & \textbf{197} & \textbf{116} & \textbf{87} & \textbf{400} \\
\bottomrule
\end{tabular}
\caption{Difficulty distribution of QuantCode-Bench tasks by source.}
\label{tab:difficulty}
\end{table}

Each task is provided in English. The code, dataset, and materials for \textbf{QuantCode-Bench} are released in the project's open repository: \url{https://github.com/LimexAILab/QuantCode-Bench}. A public benchmark page with leaderboard and resources is available at \url{https://limexailab.github.io/QuantCode-Bench/}.

\subsection{Why Backtrader}

We selected \textbf{Backtrader} because of its widespread use as an open-source framework for backtesting and prototyping trading strategies, as well as the nontrivial complexity of its API~\cite{backtrader}. Unlike simpler educational interfaces, Backtrader requires the model to correctly handle indicators, data lines, order execution methods, and indexing conventions. This makes the benchmark realistic for applied code generation and reduces the chance that success is achieved through superficial reproduction of standard templates.

\section{Evaluation Methodology}

\subsection{Validation Pipeline}

Evaluation in \textbf{QuantCode-Bench} is performed using a four-stage pipeline. A strategy is counted as successful only if it passes all stages sequentially:

\begin{enumerate}[leftmargin=1.5em]
    \item \textbf{Compilation} --- the code is syntactically correct and can be interpreted without errors.
    \item \textbf{Backtest} --- the strategy executes successfully in the evaluation environment on benchmark-provided historical market data spanning diverse assets and timeframes without runtime errors.
    \item \textbf{Trade} --- the strategy places at least one trade.
    \item \textbf{Judge} --- an LLM judge confirms that the implemented strategy matches the textual task description~\cite{zheng2023judge,gu2024surveyjudge}.
\end{enumerate}

This pipeline makes it possible to localize the failure point and decompose unsuccessful generations by level. For example, one model may achieve nearly perfect compilation but often fail at execution; another may consistently pass the backtest but generate strategies with no trades; a third may trade successfully but implement the wrong logic. This decomposition is especially important in domain-specific tasks, where a single aggregate metric obscures qualitatively different causes of failure.

\subsection{LLM Judge}

The final stage of the pipeline is designed to verify the \textbf{semantic alignment} between the generated strategy and the original task description. This stage is necessary because a strategy may be technically functional but substantively incorrect. For example, instead of an RSI-based strategy, a model might generate an SMA crossover strategy that compiles, passes the backtest, and produces trades, but does not solve the requested task.

To address this issue, we use an LLM judge that evaluates the code according to three criteria:
\begin{itemize}[leftmargin=1.5em]
    \item whether the indicators used correspond to those in the original description or to an equivalent formalization;
    \item whether the key entry, exit, and behavior logic of the strategy is implemented;
    \item whether the code constitutes a relevant implementation of the given task rather than a generic template substitution.
\end{itemize}

This approach is consistent with the broader literature on \textbf{LLM-as-a-Judge}, in which strong models are used as scalable proxies for expert evaluation in open-ended tasks~\cite{zheng2023judge,gu2024surveyjudge}.

\subsection{Evaluation Settings}

We consider two interaction settings.

In the \textbf{single-turn} setting, the model receives the task description and must generate a correct strategy on the first attempt. This scenario measures one-shot generation quality and is sensitive to the model's initial knowledge of the domain, the library, and common strategy templates.

In the \textbf{agentic multi-turn} setting, after each unsuccessful attempt the model receives structured feedback containing the error type and the corresponding system message. The model may revise the code and retry up to \textbf{10 times}. This setting measures the model's ability to iteratively repair errors, perform local search, and use diagnostic information. Similar evaluation regimes have already proven informative in broader benchmarks for code and agentic software engineering~\cite{jimenez2023swe,jain2024livecodebench,badertdinov2025swerebench}.

\section{Results}

We evaluate models of different capacities in two settings: single-turn and agentic multi-turn. Tables~\ref{tab:single_en} and~\ref{tab:agentic_en} summarize the corresponding results.

\subsection{Single-turn}

Table~\ref{tab:single_en} reports the single-turn results on \textbf{QuantCode-Bench}. The ranking already shows the central pattern of the benchmark: frontier models are almost uniformly strong on compilation, but substantially more dispersed on the later stages of the evaluation pipeline.

\begin{table}[!htbp]
\centering
\small
\begin{tabular}{lrrrr}
\toprule
\textbf{Model} & \textbf{Compilation, \%} & \textbf{Backtest, \%} & \textbf{Trade, \%} & \textbf{Judge, \%} \\
\midrule
claude-opus-4.6 & 100.0 & 98.2 & 77.2 & 75.8 \\
gpt-5.4 & 100.0 & 95.5 & 72.0 & 70.2 \\
claude-sonnet-4.5 & 100.0 & 91.5 & 71.2 & 69.8 \\
gpt-5.2-codex & 100.0 & 94.5 & 74.5 & 67.5 \\
glm-5 & 100.0 & 92.4 & 70.3 & 65.4 \\
claude-sonnet-4.6 & 100.0 & 85.8 & 66.2 & 65.0 \\
kimi-k2.5 & 99.7 & 87.3 & 67.5 & 64.8 \\
gemini-3-flash & 100.0 & 76.0 & 63.2 & 59.8 \\
grok-4.1-fast & 99.2 & 70.2 & 56.1 & 48.9 \\
deepseek-v3.2 & 100.0 & 75.8 & 50.0 & 48.8 \\
qwen3-235b & 100.0 & 72.5 & 49.0 & 48.2 \\
qwen3-coder-30b & 100.0 & 59.0 & 40.5 & 39.2 \\
gemini-2.5-flash & 99.5 & 49.2 & 33.2 & 31.2 \\
qwen3-14b & 98.0 & 42.2 & 27.8 & 25.2 \\
qwen3-8b & 99.5 & 31.8 & 19.8 & 18.5 \\
qwen3-4b & 98.8 & 24.6 & 16.4 & 12.3 \\
qwen3-1.7b & 98.1 & 23.1 & 13.7 & 7.8 \\
\bottomrule
\end{tabular}
\caption{Single-turn results on \textbf{QuantCode-Bench} (multi-timeframe).}
\label{tab:single_en}
\end{table}

As shown in Table~\ref{tab:single_en}, the single-turn results reveal a sharp divergence between the early and late stages of the pipeline. For most strong models, compilation has almost ceased to be a bottleneck. However, a high Compilation Rate does not automatically translate into a high Judge Pass. This indicates that producing a syntactically correct strategy scaffold is no longer the main limitation for modern frontier models; the major quality losses occur at the \textbf{Backtest} and \textbf{Trade} stages.

\subsection{Agentic multi-turn}

Table~\ref{tab:agentic_en} reports the cumulative performance in the agentic setting. Relative to Table~\ref{tab:single_en}, the multi-turn protocol makes it possible to observe how quickly each model converts partial failures into final task success.

\begin{table}[!htbp]
\centering
\scriptsize
\resizebox{\textwidth}{!}{%
\begin{tabular}{lrrrrrrrrr}
\toprule
\textbf{Model} & \textbf{Comp.} & \textbf{Backtest} & \textbf{Trade} & \textbf{Judge} & \textbf{AvgT} & \textbf{T1} & \textbf{T3} & \textbf{T5} & \textbf{T10} \\
\midrule
claude-opus-4.6 & 100.0 & 100.0 & 100.0 & 97.5 & 1.5 & 75.8 & 95.2 & 97.5 & 97.5 \\
claude-sonnet-4.6 & 100.0 & 99.8 & 99.8 & 96.0 & 2.0 & 65.0 & 90.2 & 93.8 & 96.0 \\
gpt-5.4 & 100.0 & 99.8 & 98.2 & 95.0 & 1.9 & 70.2 & 91.5 & 93.2 & 95.0 \\
kimi-k2.5 & 100.0 & 100.0 & 98.8 & 93.5 & 2.3 & 64.8 & 84.2 & 89.2 & 93.5 \\
claude-sonnet-4.5 & 100.0 & 100.0 & 99.5 & 93.0 & 2.0 & 69.8 & 90.0 & 91.2 & 93.0 \\
gemini-3-flash & 100.0 & 97.5 & 94.5 & 91.8 & 2.4 & 59.8 & 83.5 & 88.2 & 91.8 \\
glm-5 & 100.0 & 99.5 & 95.8 & 90.8 & 2.4 & 65.4 & 83.2 & 88.2 & 90.8 \\
gpt-5.2-codex & 100.0 & 100.0 & 99.8 & 89.8 & 2.4 & 67.5 & 84.2 & 88.2 & 89.8 \\
qwen3-235b & 100.0 & 98.2 & 93.8 & 87.2 & 3.1 & 48.2 & 74.0 & 81.2 & 87.2 \\
grok-4.1-fast & 100.0 & 97.0 & 92.2 & 84.5 & 3.2 & 48.9 & 74.5 & 79.2 & 84.5 \\
deepseek-v3.2 & 100.0 & 97.2 & 92.0 & 83.8 & 3.1 & 48.8 & 75.5 & 80.0 & 83.8 \\
qwen3-coder-30b & 100.0 & 86.2 & 76.0 & 68.0 & 4.7 & 39.2 & 57.0 & 61.8 & 68.0 \\
qwen3-14b & 100.0 & 83.2 & 67.0 & 62.7 & 5.3 & 25.2 & 51.0 & 56.0 & 62.7 \\
gemini-2.5-flash & 100.0 & 79.8 & 65.2 & 62.5 & 5.2 & 31.2 & 52.0 & 57.8 & 62.5 \\
qwen3-8b & 100.0 & 66.2 & 48.9 & 47.6 & 6.5 & 18.5 & 30.9 & 40.2 & 47.6 \\
qwen3-4b & 98.1 & 52.5 & 41.9 & 31.9 & 7.4 & 12.3 & 17.5 & 25.6 & 31.9 \\
qwen3-1.7b & 100.0 & 35.0 & 26.5 & 14.2 & 9.2 & 7.8 & 8.1 & 9.2 & 14.2 \\
\bottomrule
\end{tabular}%
}
\caption{Agentic multi-turn results on \textbf{QuantCode-Bench} (multi-timeframe). T1--T10 denote cumulative success by turn.}
\label{tab:agentic_en}
\end{table}

Taken together, the results in Table~\ref{tab:agentic_en} show that the main differences between models emerge not at the level of syntactic correctness, but at the levels of execution, generation of trading signals, and semantic compliance with the task. Iterative feedback is especially effective for strong models, for which a substantial fraction of errors are locally repairable within a small number of attempts.

\section{Error Analysis}

\subsection{Distribution by Failure Stage (single-turn)}

Table~\ref{tab:failure_stage} aggregates single-turn outcomes by the first stage at which a generation fails. This view complements the model-level results in Section~4 by showing where difficulty concentrates in the pipeline overall.

\begin{table}[!htbp]
\centering
\begin{tabular}{lr}
\toprule
\textbf{Stage} & \textbf{\%} \\
\midrule
Success & 48.2 \\
Judge rejected & 2.7 \\
No trades & 17.8 \\
Backtest fail & 26.8 \\
Compilation fail & 0.3 \\
Reasoning loops & 4.2 \\
\bottomrule
\end{tabular}
\caption{Failure stage distribution in the single-turn setting.}
\label{tab:failure_stage}
\end{table}

As summarized in Table~\ref{tab:failure_stage}, the key finding is that compilation has almost ceased to be the main problem for modern models. The main failure points lie at later stages of the pipeline, namely \textbf{Backtest} and \textbf{No trades}. This indicates that in trading-strategy generation tasks, the core difficulty is no longer Python syntax, but rather the correct operationalization of the strategy within a domain-specific execution environment.

\subsection{Classification of Backtest Errors and Late-Stage Failures}

Table~\ref{tab:error_types} provides a finer-grained taxonomy of runtime and late-stage failures. Unlike Table~\ref{tab:failure_stage}, which records only the first failed stage, this breakdown exposes the dominant technical and semantic patterns inside the broad \textbf{Backtest}, \textbf{No trades}, and late-judge failure categories.

\begingroup
\setlength{\LTleft}{0pt}
\setlength{\LTright}{0pt}
\small
\begin{longtable}{p{0.37\textwidth}rp{0.43\textwidth}}
\caption{Taxonomy of backtest and late-stage failures.}
\label{tab:error_types}\\
\toprule
\textbf{Error type} & \textbf{\%} & \textbf{Description} \\
\midrule
\endfirsthead
\caption[]{Taxonomy of backtest and late-stage failures (continued).}\\
\toprule
\textbf{Error type} & \textbf{\%} & \textbf{Description} \\
\midrule
\endhead
\midrule
\multicolumn{3}{r}{\emph{Continued on next page}}\\
\endfoot
\bottomrule
\endlastfoot
Signal conditions do not activate on data & 17.8 & Code compiles and backtests, but entry or exit conditions never trigger. \\
\texttt{\_\_bool\_\_} / Line object errors & 13.1 & Using Backtrader Line objects in boolean context without \texttt{[0]} indexing. \\
Missing attribute/method & 3.9 & Access to nonexistent attributes of the Backtrader API. \\
Wrong API params & 3.9 & Incorrect constructor arguments for indicators (e.g., \texttt{MACD} \texttt{period\_fast}). \\
Strategy doesn't match task & 2.7 & Strategy is executable and trades but does not match the specification. \\
Type/NoneType errors & 2.1 & Operations involving \texttt{None} or incompatible types. \\
Other runtime errors & 1.4 & Other runtime failures. \\
Syntax errors (runtime) & 1.2 & Syntax errors caught at runtime in \texttt{exec()}. \\
Index out of range & 1.0 & Accessing data before enough history accumulated. \\
Compilation error & 0.3 & Structural compilation errors. \\
Execution timeout & 0.2 & Infinite loops or time limit exceeded. \\
\end{longtable}
\endgroup

As shown in Table~\ref{tab:error_types}, the most frequent failure type involves strategies that compile successfully and pass the backtest, but place no trades on the data. This is typically caused by overly strict or unrealistic entry conditions, insufficient historical context for feature computation, or incorrect operationalization of indicator logic. The second most frequent category is \textbf{\texttt{\_\_bool\_\_} / Line object errors}, which reflects incorrect handling of Backtrader line objects in boolean conditions. \textbf{Missing attribute/method} errors account for a smaller share of failures than the leading categories, indicating that direct API hallucinations are less prevalent than logic-activation and line-object handling failures.

\subsection{Errors in the Agentic Setting}

Tables~\ref{tab:agentic_outcomes} and~\ref{tab:agentic_shift} summarize the final outcome distribution in the agentic setting and compare selected single-turn error categories with the last-turn composition for failed agentic trajectories.

\begin{table}[!htbp]
\centering
\begin{tabular}{lr}
\toprule
\textbf{Outcome} & \textbf{\%} \\
\midrule
Success & 81.8 \\
Judge rejected & 5.6 \\
No trades & 4.7 \\
Backtest fail & 7.9 \\
\bottomrule
\end{tabular}
\caption{Final outcome distribution in the agentic setting.}
\label{tab:agentic_outcomes}
\end{table}

\begin{table}[!htbp]
\centering
\small
\resizebox{\textwidth}{!}{%
\begin{tabular}{lrrl}
\toprule
\textbf{Error type} & \textbf{Turn 1} & \textbf{Last turn} & \textbf{Trend} \\
\midrule
Signal conditions do not activate on data & 17.8\% & 23.7\% & grows \\
\texttt{\_\_bool\_\_} / Line object errors & 13.1\% & 28.9\% & grows --- hardest to fix iteratively \\
Missing attribute/method & 3.9\% & 3.3\% & slight drop \\
Wrong API params & 3.9\% & 6.6\% & grows \\
Strategy doesn't match task & 2.7\% & 23.7\% & grows --- semantic failures accumulate \\
Compilation fail & 0.3\% & 6.2\% & grows \\
\bottomrule
\end{tabular}%
}
\caption{Comparison of selected error categories in the single-turn setting and at the last turn for failed agentic trajectories. The Turn 1 column reports the corresponding single-turn shares from Table~\ref{tab:error_types}.}
\label{tab:agentic_shift}
\end{table}

Compared with the single-turn profile, the composition of unresolved failures at the last agentic turn shifts toward categories that reflect persistent semantic and logic-level problems. In particular, the shares of \textbf{Strategy doesn't match task}, \textbf{Signal conditions do not activate on data}, and \textbf{\texttt{\_\_bool\_\_} / Line object errors} are larger among the remaining failures at the last turn, whereas \textbf{Missing attribute/method} remains comparatively infrequent. For strategies that remain unsolved after 10 attempts, Judge rejection becomes one of the main causes of final failure.

This shows that iterative debugging is effective primarily for repairing technical errors, but substantially less effective when the model misunderstands the task itself. Therefore, the agentic setting mainly addresses program-repair problems, but does not fully eliminate limitations in the semantic interpretation of natural-language specifications.

\FloatBarrier

\section{Discussion}

The generation of algorithmic trading strategies constitutes a distinct class of tasks at the intersection of programming, financial logic, and agentic search. The results of \textbf{QuantCode-Bench} show that even very strong models have already mastered some components of this problem, while remaining significantly limited in others.

The first important conclusion is that modern LLMs have largely solved the problem of \textbf{surface-level syntactic generation}. For strong models, compilation is no longer a meaningful bottleneck. The main challenge has shifted to the level of \textbf{operational formalization}: the model must not merely express a trading idea in code, but do so in a way that is executable, activatable on data, and semantically correct. This shift is particularly important for understanding what kinds of benchmarks are needed for the next stage of code-generation evaluation.

The second conclusion concerns the contrast between \textbf{single-turn} and \textbf{agentic} settings. The large improvement observed under iterative feedback shows that a substantial fraction of errors in \textbf{QuantCode-Bench} belongs to the class of locally repairable specification or API violations rather than to a fundamental inability to generate a strategy at all. This means that the practical usefulness of a model in this setting is determined not only by its single-turn accuracy, but also by its effectiveness in iterative code repair. This conclusion is conceptually aligned with observations from \textbf{SWE-Bench Verified} and \textbf{SWE-rebench}, where interactivity and repair play a major role in realistic evaluation of model capability~\cite{jimenez2023swe,badertdinov2025swerebench}.

The third conclusion concerns the nature of difficulty. In benchmarks of the natural-language-to-code or natural-language-to-strategy type, difficulty is not simply a function of conceptual depth. \textbf{Specification quality} plays a major role. Vague and conversational descriptions are often harder for models than more formal and parameterized formulations, even when the latter are conceptually more complex.

The fourth conclusion concerns the distinction between general-purpose models and code-specialized models. The results of \textbf{QuantCode-Bench} show that specialization in programming does not guarantee superiority in domain-specific strategy generation. The likely reason is that the task requires not only technical discipline, but also precise interpretation of financial intent, translation of textual descriptions into behaviorally meaningful logic, and selection of realistic trigger conditions. In this setting, general-purpose models with stronger semantic and instruction-following capabilities often outperform specialized coding models.

The fifth conclusion concerns the role of the judge. Without semantic validation, Trade Rate systematically overestimates true success, because some strategies pass all technical checks yet still do not match the task. This is particularly important for open-ended benchmarks, where a model may generate a technically correct but semantically irrelevant implementation instead of the requested strategy. This is also supported by our reinforcement-learning experiments: when the reward function includes only technical pipeline completion and the presence of a trade, the model tends to exploit the reward by repeatedly generating the same working template that trades on the data but is unrelated to the original task. Adding the Judge stage eliminates this behavior by making semantic compliance part of the reward objective. In this sense, \textbf{QuantCode-Bench} shows that for open-ended domain-specific tasks, technical metrics alone are insufficient, and semantic validation must be part of the main evaluation procedure~\cite{zheng2023judge,gu2024surveyjudge}.

\section{Limitations}

Although \textbf{QuantCode-Bench} covers an important and practically relevant class of tasks, the current version of the benchmark has several limitations.

First, all strategies are evaluated within a single framework and a single execution environment, namely \textbf{Backtrader}~\cite{backtrader}. This improves experimental control and reduces infrastructural variability, but also limits the transferability of the results to other algorithmic-trading libraries and environments. Natural directions for extending the benchmark include \textbf{QuantConnect/LEAN} and \textbf{Zipline}~\cite{quantconnect,zipline}. Evaluation across multiple frameworks would make it possible to better disentangle a model's capacity for domain-specific strategy synthesis from its adaptation to a particular API.

Second, the final semantic evaluation relies on an \textbf{LLM judge}. Although this approach substantially strengthens the benchmark relative to purely technical validation, it does not provide an absolute guarantee of semantic correctness. The judge may overlook subtle mismatches, especially in cases where the strategy partially aligns with the task but diverges in details of the logic. In addition, the usual concerns associated with LLM-as-a-Judge remain relevant, including positional, stylistic, and model-specific biases that may influence the final evaluation~\cite{zheng2023judge,gu2024surveyjudge}.

Third, the benchmark does not evaluate \textbf{profitability}, risk robustness, or the economic quality of the generated strategy. The presence of trades and alignment with the text do not imply that a strategy is effective as a trading system. In the present work, our focus is specifically on the ability of models to generate executable strategies from descriptions, not on the investment quality of those strategies.

\section{Conclusion}

In this work, we introduced \textbf{QuantCode-Bench}, a benchmark for evaluating the ability of large language models to generate executable algorithmic trading strategies. The benchmark formalizes the task as a sequence of nested requirements: syntactic correctness, successful execution, the presence of trades, and semantic alignment with the original description. This structure makes it possible to evaluate not only surface-level code quality, but also the deeper ability of a model to translate a natural-language trading idea into a behaviorally valid implementation.

The results show that even frontier models remain far from fully solving the task in the one-shot setting: the maximum single-turn Judge Pass now reaches roughly three quarters of the benchmark, but still falls well short of saturation. At the same time, the agentic setting with iterative feedback yields a sharp and consistent improvement, raising the best models to \textbf{95--98\%}. This indicates that a substantial fraction of errors is repairable and that model behavior in an interactive debugging loop may be at least as important as its accuracy in single-turn generation.

Taken together, the results of \textbf{QuantCode-Bench} show that trading-strategy generation requires simultaneous command of a specialized API, the ability to construct executable code, the capacity to formulate realistic trading logic, and adherence to the semantics of a natural-language specification. Modern models already perform well on the syntactic and basic infrastructural layers of the task, but still exhibit limitations in robust one-shot formalization of trading intent and in the precise implementation of the requested strategy.

\textbf{QuantCode-Bench} can serve as a useful tool for future research in domain-specific code generation, agentic software repair, and evaluation of LLMs in the financial domain.

\clearpage
\appendix
\section*{Appendix}
\section{Example Tasks from QuantCode-Bench}

Below we present one representative example for each difficulty level in \textbf{QuantCode-Bench}.

\needspace{11\baselineskip}
\subsection{Easy example}

\begin{taskbox}
\footnotesize
\textbf{Source:} StackExchange \hfill \textbf{Difficulty:} Easy

\vspace{0.25em}
\textbf{Task:}

Calendar-based monthly strategy for \texttt{SPY} on daily data over a five-year backtest window.

\vspace{0.25em}
\textbf{Indicator setup:}
\begin{itemize}[itemsep=1pt, topsep=1pt]
    \item No technical indicators are used.
    \item Signals are triggered only by the calendar event marking the first trading day of each month.
\end{itemize}

\textbf{Entry rule:}
\begin{itemize}[itemsep=1pt, topsep=1pt]
    \item Open a long position on the first trading day of each month, regardless of price action.
\end{itemize}

\textbf{Exit rule:}
\begin{itemize}[itemsep=1pt, topsep=1pt]
    \item Close all open positions on the last trading day of the month.
\end{itemize}

\textbf{Risk management:}
\begin{itemize}[itemsep=1pt, topsep=1pt]
    \item Position size is 100\% of available capital with a fixed initial capital of \$1000.
    \item The strategy reports the initial and final portfolio value after backtesting.
\end{itemize}
\end{taskbox}

\needspace{16\baselineskip}
\subsection{Medium example}

\begin{taskbox}
\footnotesize
\textbf{Source:} Reddit \hfill \textbf{Difficulty:} Medium

\vspace{0.25em}
\textbf{Task:}

\texttt{SOFI} mean reversion with intraday pullback or flag-pattern entries on the 15-minute timeframe.

\vspace{0.25em}
\textbf{Indicator setup:}
\begin{itemize}[itemsep=1pt, topsep=1pt]
    \item Use \texttt{EMA(10)} and \texttt{EMA(20)} as short- and medium-term trend filters.
    \item Use a volatility-sensitive filter based on \texttt{VRVP}.
    \item Restrict entries to cases where price trades within 1\% of the 52-week high.
\end{itemize}

\textbf{Entry rules:}
\begin{itemize}[itemsep=1pt, topsep=1pt]
    \item Enter long on a pullback to a gap-fill level or to a rising intraday EMA, followed by an upward reversal.
    \item Alternatively, skip the initial move and enter on a breakout from a 15-minute flag pattern in the direction of the main trend.
\end{itemize}

\textbf{Exit rules:}
\begin{itemize}[itemsep=1pt, topsep=1pt]
    \item Close the position at a local stop-loss level or with a default stop of roughly 1--2\% from entry.
    \item Take profit at the 52-week high or at a default target of roughly 3--5\%.
    \item Exit if price closes the day below the 10- or 20-day EMA.
\end{itemize}

\textbf{Risk management and filters:}
\begin{itemize}[itemsep=1pt, topsep=1pt]
    \item Size the trade so that risk per position remains within 1--2\% of total capital.
    \item Avoid chasing the initial gap move; prefer asymmetric entries after retracement or consolidation.
\end{itemize}
\end{taskbox}

\needspace{16\baselineskip}
\subsection{Hard example}

\begin{taskbox}
\footnotesize
\textbf{Source:} GitHub \hfill \textbf{Difficulty:} Hard

\vspace{0.25em}
\textbf{Task:}

One-minute scalping strategy for \texttt{SNAP} using momentum, liquidity, and volatility filters.

\vspace{0.25em}
\textbf{Indicator construction:}
\begin{itemize}[itemsep=1pt, topsep=1pt]
    \item Use \texttt{Stochastic(5,3,3)} as a proxy for buyer/seller pressure and \texttt{CrossOver(\%K, \%D)} as the trigger.
    \item Use \texttt{RSI(3)} for micro-momentum confirmation.
    \item Use \texttt{SMA(volume, 20)} as a liquidity filter and \texttt{ATR(14)} with a threshold of 0.03 as a volatility filter.
\end{itemize}

\textbf{Entry logic:}
\begin{itemize}[itemsep=1pt, topsep=1pt]
    \item Enter long when \texttt{\%K} crosses above \texttt{\%D}, \texttt{\%K > 80}, \texttt{RSI(3) > 55}, current volume exceeds \(1.2 \times \texttt{SMA(volume, 20)}\), and \texttt{ATR(14) \(\geq\) 0.03}.
    \item Enter short under the mirrored conditions: downward crossover, \texttt{\%K < 20}, \texttt{RSI(3) < 45}, sufficient volume, and the same ATR filter.
\end{itemize}

\textbf{Exit logic:}
\begin{itemize}[itemsep=1pt, topsep=1pt]
    \item Use a take-profit of \$0.01 from entry and a stop-loss of \$0.02.
    \item Close on signal reversal or after a timeout of three bars.
\end{itemize}

\textbf{Implementation constraints:}
\begin{itemize}[itemsep=1pt, topsep=1pt]
    \item Limit position size to 100--500 shares, with 100 shares as the default.
    \item Trade only during the main session, skip the first three minutes, and require sufficient rolling volume; an optional \$25{,}000 balance filter may be applied for PDT compliance.
\end{itemize}
\end{taskbox}


\begin{thebibliography}{10}

\bibitem{badertdinov2025swerebench}
Ibragim Badertdinov et~al.
\newblock Swe-rebench: An automated pipeline for task collection and
  decontaminated evaluation of software engineering agents, 2025.

\bibitem{bigeard2025financeagent}
Antoine Bigeard et~al.
\newblock Finance agent benchmark: Benchmarking llms on real-world financial
  research tasks, 2025.

\bibitem{choi2025finagentbench}
Chanyeol Choi et~al.
\newblock Finagentbench: A benchmark dataset for agentic retrieval in financial
  question answering, 2025.

\bibitem{gu2024surveyjudge}
Jiawei Gu et~al.
\newblock A survey on {LLM}-as-a-judge, 2024.

\bibitem{islam2023financebench}
Pranab Islam et~al.
\newblock Financebench: A new benchmark for financial question answering, 2023.

\bibitem{jain2024livecodebench}
Naman Jain et~al.
\newblock Livecodebench: Holistic and contamination-free evaluation of large
  language models for code, 2024.

\bibitem{jimenez2023swe}
Carlos~E. Jimenez et~al.
\newblock Swe-bench: Can language models resolve real-world github issues?,
  2023.

\bibitem{shi2025finr1}
Zhaowei Liu et~al.
\newblock Fin-r1: A large language model for financial reasoning through
  reinforcement learning, 2025.

\bibitem{merrill2026terminalbench}
Miles~A. Merrill et~al.
\newblock Terminal-bench: Benchmarking agents on hard, realistic tasks, 2026.

\bibitem{shah2025fino1}
Lingfei Qian et~al.
\newblock Fino1: On the transferability of reasoning-enhanced llms and
  reinforcement learning to finance, 2025.

\bibitem{quantconnect}
{QuantConnect}.
\newblock Quantconnect {LEAN} documentation.
\newblock \url{https://www.quantconnect.com/docs/}, 2026.

\bibitem{backtrader}
Daniel Rodriguez.
\newblock Backtrader.
\newblock \url{https://www.backtrader.com/}, 2015.

\bibitem{xie2023pixiu}
Qianqian Xie et~al.
\newblock Pixiu: A large language model, instruction data and evaluation
  benchmark for finance, 2023.

\bibitem{xie2024finben}
Qianqian Xie et~al.
\newblock Finben: A holistic financial benchmark for large language models,
  2024.

\bibitem{zheng2023judge}
Lianmin Zheng et~al.
\newblock Judging {LLM}-as-a-judge with {MT}-bench and chatbot arena, 2023.

\bibitem{zipline}
{Zipline}.
\newblock Zipline documentation.
\newblock \url{https://zipline.ml4trading.io/}, 2026.

\end{thebibliography}
\end{document}